\documentclass[sigconf]{acmart}

\usepackage{booktabs} % For formal tables
\usepackage{tikz}
\usetikzlibrary{arrows.meta,positioning,calc,shapes,shapes.geometric,decorations.pathreplacing}

\setcopyright{acmcopyright}
\acmYear{2025}
\copyrightyear{2025}

\setcopyright{acmlicensed}\acmConference[RACS '25]{International Conference on Research in Adaptive and Convergent Systems}{November 16--19, 2025}{Ho Chi Minh, Vietnam}
\acmBooktitle{International Conference on Research in Adaptive and Convergent Systems (RACS '25), November 16--19, 2025, Ho Chi Minh, Vietnam}
\acmDOI{10.1145/3769002.3769956}
\acmISBN{979-8-4007-2231-8/2025/11}

\acmArticle{4}
\acmPrice{15.00}

\usepackage{multirow}
\usepackage{subcaption}
\usepackage{amsmath}
\usepackage{mathtools}
\usepackage{amsthm}
\usepackage{pifont}% http://ctan.org/pkg/pifont
\newcommand{\cmark}{\ding{51}}%
\newcommand{\xmark}{\ding{55}}%

\begin{document}
\title{Discovering Spatial Correlations of Earth Observations in Global Atmospheric State Estimation by using Adaptive Graph Structure Learning}
\title{Adaptive Graph Structure Learning for Weather Prediction by Discovering Spatial Correlations of Earth Observations}
\title{Discovering Spatial Correlations of Earth Observations \\ for Weather Forecasting by using Graph Structure Learning}
%\title{Dynamic Graph Structure Learning for Weather Forecasting \\ with Heterogeneous Earth Observations}
%\titlenote{Produces the permission block, and
%  copyright information}
%\subtitle{Extended Abstract}
%\subtitlenote{The full version of the author's guide is available as
%  \texttt{acmart.pdf} document}
  
\renewcommand{\shorttitle}{Discovering Spatial Correlations of Earth Observations for Weather Forecasting %using Graph Structure Learning
}

\author{Hyeon-Ju Jeon}
%\authornote{Dr.~Trovato insisted his name be first.}
\orcid{1234-5678-9012}
\affiliation{%
  \institution{Data Assimilation Group\\Korea Institute of Atmospheric Prediction Systems}
  \streetaddress{35, 5-gil, Boramae-ro}
  \city{Seoul}
  \country{Republic of Korea}
  \postcode{07071}  
}
\email{hjjeon@kiaps.org}

\author{Jeon-Ho Kang}
%\authornote{Dr.~Trovato insisted his name be first.}
\affiliation{%
  \institution{Data Assimilation Group\\Korea Institute of Atmospheric Prediction Systems}
  \streetaddress{35, 5-gil, Boramae-ro}
  \city{Seoul}
  \country{Republic of Korea}
  \postcode{07071}  
}
\email{jhkang@kiaps.org}

\author{In-Hyuk Kwon}
%\authornote{Dr.~Trovato insisted his name be first.}
\affiliation{%
  \institution{Data Assimilation Group\\Korea Institute of Atmospheric Prediction Systems}
  \streetaddress{35, 5-gil, Boramae-ro}
  \city{Seoul}
  \country{Republic of Korea}
  \postcode{07071}  
}
\email{ihkwon@kiaps.org}

\author{O-Joun Lee}
\authornote{Correspondence to: \texttt{ojlee@catholic.ac.kr}; Tel.: +82-2-2164-5516}
\orcid{0000-0001-8921-5443}
\affiliation{%
  \institution{Network Science Lab\\The Catholic University of Korea}
  \streetaddress{43 Jibong-ro}
  \city{Bucheon} 
  \country{Republic of Korea}
  \postcode{14662} 
}
\email{ojlee@catholic.ac.kr}

\renewcommand{\shortauthors}{H.-J. Jeon et al.}

\begin{abstract}
This study aims to improve the accuracy of weather predictions by discovering spatial correlations between Earth observations and atmospheric states.
%This study aims to discover spatial correlations between Earth observations and atmospheric states to improve the forecasting accuracy of global atmospheric state estimation, which are usually conducted using conventional numerical weather prediction (NWP) systems and is the beginning of weather forecasting. 
Existing numerical weather prediction (NWP) systems predict future atmospheric states at fixed locations, which are called NWP grid points, by analyzing previous atmospheric states and newly acquired Earth observations.% without fixed locations.
However, the shifting locations of observations and the surrounding meteorological context induce complex, dynamic spatial correlations that are difficult for traditional NWP systems to capture, since they rely on strict statistical and physical formulations.
%Thus, the surrounding meteorological context and the changing locations of the observations make spatial correlations between atmospheric states and observations over time. 
To handle complicated spatial correlations, which change dynamically, we employ a spatiotemporal graph neural networks (STGNNs) with structure learning. 
However, structure learning has an inherent limitation that this can cause structural information loss and over-smoothing problem by generating excessive edges. 
To solve this problem, we regulate edge sampling by adaptively determining node degrees and considering the spatial distances between NWP grid points and observations. 
We validated the effectiveness of the proposed method (CloudNine-v2) using real-world atmospheric state and observation data from East Asia, achieving up to 15\% reductions in RMSE over existing STGNN models. 
Even in areas with high atmospheric variability, CloudNine-v2 consistently outperformed baselines with and without structure learning.
\end{abstract}




\maketitle

\section{Introduction} \label{Sec:intro}
Accurate weather prediction is critical for disaster prevention, resource management, and public safety, with a wide impact in many domains \cite{Zhou_2021,Wu2024}.
Traditional numerical weather prediction (NWP) systems rely on data assimilation methods \cite{Kwon2018,Bonavita2015,Clayton2012} to combine observational data and physical model states. 
However, these methods are computationally intensive and require extensive pre-processing and quality control to handle inconsistencies, missing values, and heterogeneous formats. 
Additionally, when observational information is projected onto the model’s initial state, much of the raw measurements signal can be distorted or lost, particularly when the model's resolution or physical assumptions deviate from real-world conditions.

\begin{figure*}
\centering
  \begin{subfigure}[h]{0.2\linewidth}
    \centering
    \includegraphics[width=\linewidth]{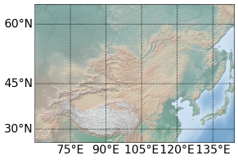}
    \caption{Topography of East Asia.}
    \label{fig:var_node_stats}
   \end{subfigure}
   \hfill
  \begin{subfigure}[h]{0.78\linewidth}
      \begin{subfigure}{\linewidth}
      \centering
      \includegraphics[width=\linewidth]{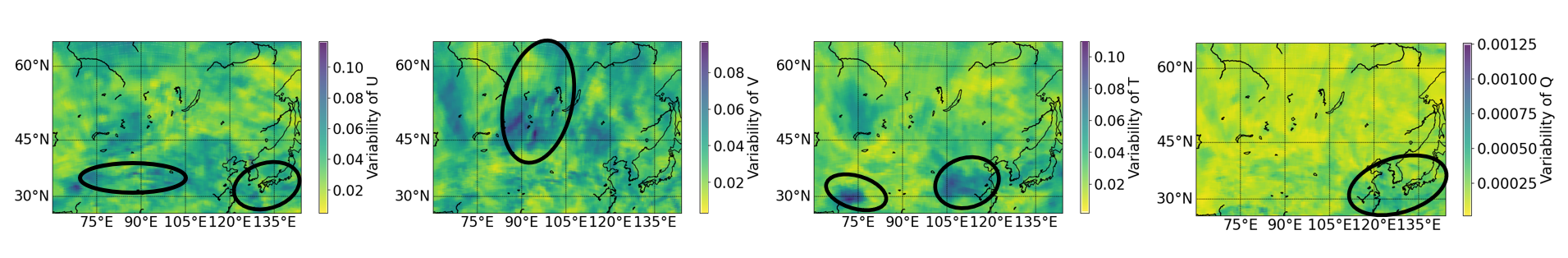}
      \caption{Spatial distribution of nodes characterized by atmospheric variability.}
      \label{fig:motiv_map}
      \end{subfigure}
      
      \begin{subfigure}{\linewidth}
      \centering
      \includegraphics[width=\linewidth]{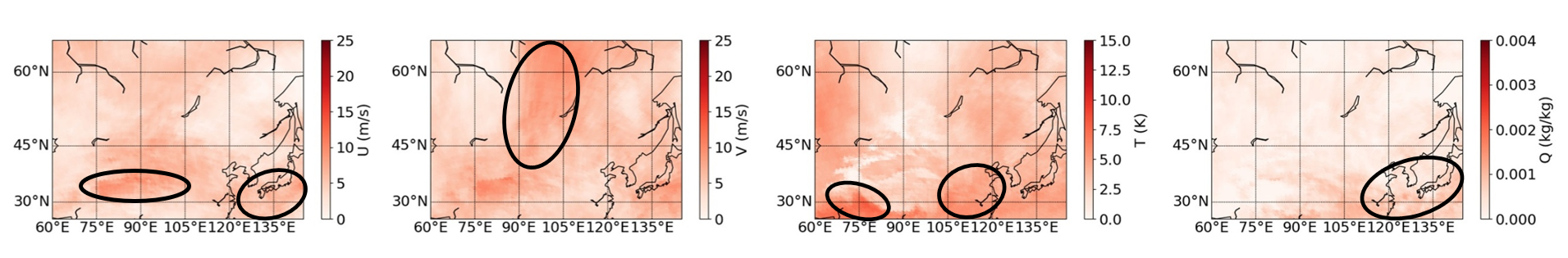}
      \caption{Spatial distribution of node-level RMSE of atmospheric state estimation by CloudNine \cite{Jeon24MLST}.}
      \label{fig:motiv_mae}
  \end{subfigure}
  \end{subfigure}
%}
\caption{
  Regional variability and its impact on atmospheric state estimation errors. Areas with higher variability (black eclipses) tend to show larger RMSE values. 
}
  \label{fig:motiv_combined}
\end{figure*}

Graph neural networks (GNNs) have emerged as a powerful framework for representing unstructured and relational data in various scientific domains.
In meteorological applications, GNNs have been used to model spatial dependencies among heterogeneously distributed ground-based observation stations \cite{Jeon_2022_solar,Jeon2024}, to correct error in NWP systems \cite{Wu2024}, and to quantify the impact of observations on NWP grid points \cite{Jeon24MLST}.
%They have also been used for tasks such as NWP error correction \cite{Wu2024} and determining the impact of observations on the analysis state \cite{Jeon24MLST}.
Spatiotemporal GNNs (STGNNs) further incorporate temporal dynamics, enabling sequence-to-sequence prediction of evolving physical systems \cite{Lam2023, Li2018}. 
However, most existing STGNN variants rely on a pre-defined static graph structure across all time steps, implicitly assuming a fixed observation topology.
This assumption is particularly limiting in meteorological settings, where the coverage and sensor footprints of observation platforms, such as polar-orbiting or geostationary satellites, change dynamically over time. 
Consequently, it becomes difficult to exploit original observations without information loss when the observation configuration varies over time.
%This assumption poses a particular challenge in meteorological networks due to the diversity of observation platforms (e.g., ground stations, ships, and various satellite types), differences in measurement frequency and spatial coverage, and region-specific atmospheric variability.
%These factors demand a flexible, context-aware representation. 
%This assumption is problematic in scenarios involving observation platforms, such as polar-orbiting or geostationary satellites, whose coverage and sensor footprints change dynamically over time. 
Although some recent studies \cite{Wang_2025_stgnn,Wu_gwnet} have explored dynamic graph modeling based on spatial proximity between observations, most approaches still depend on heuristically defined graphs. 
Such static representations cannot capture the broad and dynamically changing influence radii characteristic of regions with high atmospheric variability, such as coastal or mountainous areas. 
As illustrated in Figure~\ref{fig:motiv_combined}, prediction errors in these regions are often significantly higher.
%In particular, heuristically defined static graphs cannot capture the broad, dynamically shifting influence radii that characterize regions with high atmospheric variability, such as coastal or mountainous areas.
%Consequently, prediction errors are often higher in these regions, as illustrated in Figure~\ref{fig:motiv_combined}. 
%Mean absolute errors (MAEs) across several meteorological variables are consistently greater in regions with high variability compared to regions with low variability. 
Our empirical evaluation confirms that models relying on fixed adjacency matrices incur up to 30\% higher RMSE in high‐variability regions compared to low‐variability regions, underscoring their inability to capture dynamic local interactions.
%These observations underscore the need for graph learning methods that can adaptively model context-specific, time-varying spatial relationships to improve predictive performance in complex real-world weather conditions.

Recent work on graph structure learning \cite{Jiang_2023_megaCRN} has sought to better capture dynamic, context-dependent interactions in spatiotemporal data. 
However, meteorological observations are inherently heterogeneous, spanning ground stations, radiosondes, ships, and diverse satellite sensors, each with distinct spatial distributions, sampling frequencies, and measured variables. 
Their spatial coverage and temporal availability fluctuate dramatically due to orbital patterns, sensor outages, and meteorological conditions.
Furthermore, regional atmospheric variability, particularly in complex terrain such as coastal or mountainous areas, underscores the need for a graph structure that can adapt dynamically.

To address these issues, this study aims to extend existing graph structure learning method, which often assumes node homogeneity and temporal stationarity, by proposing a novel framework capable of capturing the intrinsic characteristics of meteorological networks, such as multi-source heterogeneity, dynamically evolving spatial relationships, and localized atmospheric variability.
Recent work on dynamic graph learning \cite{Wu_gwnet,Wang_2025_stgnn} has primarily relied on spatial proximity to construct time-varying graphs, but these approaches overlook the heterogeneous nature of meteorological platforms and the highly localized influence of atmospheric variability.
Our framework dynamically infers regional adjacency matrices for each prediction target (grid point) at every time step using observation features and metadata (e.g., geographic coordinates and sensor types).
%We propose an adaptive graph-structured learning framework that dynamically infers regional adjacency matrices from observation features (i.e., sensor measurements) and metadata such as geographic coordinates and sensor attributes.
%Unlike traditional methods, which rely on fixed or heuristically defined graph topologies, our approach dynamically infers a regional adjacency matrix for each prediction target (grid point) at every time step. 
%This process utilizes observations, as well as metadata, including geographic coordinates and sensor types.
By constructing a $k$-hop subgraph around each grid point, we integrate NWP state vectors with heterogeneous observations from multiple platforms. 
Node-type-specific encoders map diverse inputs into a unified embedding space, enabling fair comparison across sensing modalities.
%Node features are projected into a unified embedding space through node-type-specific encoders. 
Then, we use a differentiable Gumbel-Softmax mechanism to adaptively select the $k$ most relevant neighbors for each node, balancing feature similarity and spatial proximity.
The design mitigates spurious connections, avoids over-smoothing, while preserving meaningful local structures.
These dynamically inferred adjacency matrices are integrated into a STGNN, where a GNN-based encoder captures spatial dependencies and a GRU-based decoder aggregates temporal information over a sliding window. 
This design enables our model to flexibly adapt to local atmospheric variability and rapidly changing observation topologies.
To our knowledge, this is the first framework that jointly models heterogeneous observations, dynamic spatial relationships, and localized variability in a unified STGNN architecture.
Experiments confirm that our method achieves robust and accurate forecasts, significantly improving performance in high-variability regions without sacrificing generalization elsewhere.

%As a result, the proposed model allows us to accommodate both feature similarities and graph structure, achieving significant improvements in high-variation regions without sacrificing generalization elsewhere.
%Our experimental results validate that the proposed model achieves accurate and robust integration of diverse observational data, leading to enhanced predictive performance in spatiotemporal weather forecasting.

The main contributions of this work are as follows:
\begin{itemize}
\item We propose a novel, adaptive framework for learning graph structures that can dynamically infer regional adjacency matrices for each prediction target. This framework integrates observation features and metadata to enable context-aware connectivity in meteorological networks.
\item We introduce a differentiable Gumbel-Softmax-based edge selection mechanism that identifies the most relevant neighboring nodes for each grid point, allowing the model to flexibly capture both feature similarity and spatial relationships.
\item We have developed a scalable pipeline for constructing subgraphs and encoding spatiotemporal data. This enables the efficient integration of multi-source, heterogeneous observations and NWP model states within a unified STGNN architecture.
\item Extensive experimentation using real-world meteorological datasets has demonstrated that our method significantly improves forecasting accuracy, particularly in regions with high atmospheric variability, while maintaining robust performance across diverse conditions.
\end{itemize}

\section{Related Work} \label{Sec:related}

Various studies \cite{Jeon_2022_solar, Jeon24MLST, Jeon2024} attempted to apply STGNNs to analyze complicated spatial correlations between meteorological variables. 
However, they have not paid attention to discover spatial correlations beyond given graph topologies by employing structure learning. 
Although \citet{Chen2024} applied adaptive weights to adjacency matrices, this approach only can reduce noisy correlations, not discovering missing correlations. 

Thus, this section mainly introduces existing studies that apply structure learning to STGNNs. 
\citet{Wang_2025_stgnn} exhibited one of widely-used approaches of structure learning that calculate feature correlations between nodes by dot products of node embeddings and normalize them with softmax function. 
Graph WaveNet \cite{Wu2019} and MegaCRN \cite{Jiang_2023_megaCRN} used a similar structure learning method, and StemGNN \cite{Cao_2020_stemgnn} added linear transformations for node features and a scaling factor, similar to self-attention. 
SLCNN \cite{Zhang_2020_slcnn} used two adjacency matrices. 
One was directly updated by backward propagation, and the other was obtained by dot products of node embeddings and learnable parameters. 
With a similar approach, \citet{Ta2022} added missing correlations to orginal adjacency matrices. 

However, structure learning inherently causes loss of structural information, and these methods have limitations in their absence of mechanisms for regulating changes in original graph topologies. 
For general GNNs, \citet{Qian_2024_rewire, Qian_2024_rewire2} sample top-$k$ edges for each node according to node feature correlations. 
\citet{Saha_2023} added a step for estimating node degrees to this method to adaptively set the number of neighbors $k$. 
FoSR \cite{Karhadkar_2023} focused on resolving over-squashing problem while preventing over-smoothing problem and structural information loss by adding edges that minimize degree-scaled dot products between node embeddings. 
\citet{Ye_2023} directly updated masks for adjacency matrices, but used a regularization term for the masks.
For continuous-time dynamic graphs, \citet{Zhang_2023} employed top-$k$ edge selection. 
They also considered temporal distances between nodes in assessing edge candidates and replaced Softmax for normalizing node correlations with Gumbel-Softmax \cite{Jang_2017_gumbel}, which can remove edges with low node correlations. 
Since NWP systems handle discrete time points, we do not consider temporal distances, but we use spatial distances not to generate unrealistic edges and employ the adaptive top-$k$ edge sampling \cite{Zhang_2023} with Gumbel-Softmax too. 

\section{Method}

\begin{figure*}[t]
    \includegraphics[width=\linewidth]{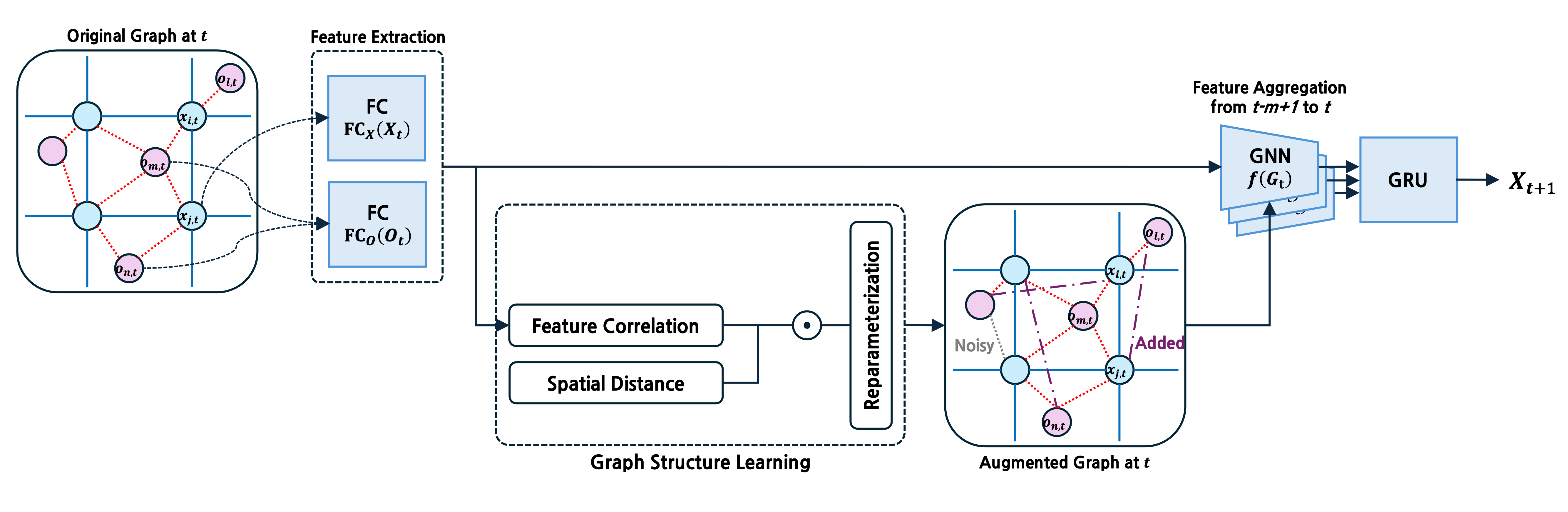}
    \caption{An illustration of the overall process of CloudNine-v2. The model consists of three stages: (1) extracting features in both NWP grid points and multi-source observations, (2) adaptive graph structure learning, and (3) spatial and temporal feature extraction for weather forecasting.}
    \label{fig:method}
\end{figure*}

The NWP systems predict future atmospheric states at fixed spatial locations, which are called NWP grid points, by analyzing previous atmospheric states and Earth observations. 
We formulate our problem as an atmospheric state forecasting task as follows:
\begin{equation}
    [X_{t-m+1}, \cdots, X_t, O_{t-m+1}, \cdots, O_{t}] \xrightarrow[\theta]{F(\cdot)} X_{t+1},
\end{equation}
where $X_t \in \mathbb{R}^{\lvert X_t \rvert \times C}$ and $O_t \in \mathbb{R}^{\lvert O_t \rvert \times C}$ are NWP grid points and observations on time $t$, $C$ refers to the number of the variables, and $m$ indicates the time window size. 

%The atmospheric states consistently appear at every time point as well as their spatial locations, while observations have irregular spatial locations and a variety of variables and appear at only one time point. 
Unlike grid points, which are consistently available at fixed spatial locations, observations are irregular in both space and variable types, and often appear only once in time.
Thus, in predicting future atmospheric states, it is significant to discover which observations are correlated to the corresponding NWP grid points. 
In addition, the spatial correlations between the NWP grid points can be affected by the terrain between them, and the correlations can change over time. 
Existing studies represented the dynamic correlations as a dynamic graph $G_t=\langle V_t \ni v_{i,t}, E \ni e_{i,j,t} \rangle$, which has NWP grid points and observations $V_t = V_{Xt} \cup V_{Ot}$ as nodes and their spatial correlations as edges. 
However, most of these studies merely determined correlations between NWP grid points and observations based only on their spatial distances \cite{Jeon24MLST}. 
To resolve this issue, we attempted to employ graph structure learning at each time point to capture dynamic changes in the spatial correlations. 
Before applying the graph structure learning, the edges are initially set based on spatial distances by connecting nodes within a 50 km radius, which corresponds to the horizontal resolution of the NWP initial conditions. 
An overview of the process is depicted in Figure \ref{fig:method}.

\subsection{Subgraph Sampling}

The NWP grid is defined at a horizontal resolution of 50 km with 91 vertical levels, and the number of grid points is more than 190 thousands. 
Considering the time window and inclusion of observations, the entire graph becomes massive, and it is difficult to consider global atmospheric states and observations in predicting future atmospheric states of each grid point. 
Therefore, we employ $k$-hop subgraph sampling. 
Although atmospheric states of close locations are correlated with each other, the limited time window limits the range of influence propagation. 
Thus, if we set enough $k$-hop radius around a target node, we can aggregate enough information from $k$-hop subgraph rooted in the target node to achieve high prediction accuracy while guaranteeing scalability to handle massive meteorological data. 
A sampled subgraph rooted in a target node $v_i$ can be formulated as: 
\begin{align}
{G}(v_i) &= \left\langle {G_{1}}(v_i), {G_{2}}(v_i), \cdots, {G_{t}}(v_i) \right\rangle, 
\\
{G_{t}}(v_i) &= \left\langle {V_{t}}(v_i) \subset V_t, {E_{t}}(v_i) \subset E_t \right\rangle,
\\
{V_{t}}(v_i) &= \left\{ v_{j,t} \lvert \text{SPD}(v_{i,t},v_{j,t}) \leq k, v_{i,t},v_{j,t} \in V_{t} \right\},
\end{align}
where $\text{SPD}(\cdot,\cdot)$ indicates the shortest path distance between two nodes. 
${E_{t}}(v_i)$ include every edge between nodes in ${V_{t}}(v_i)$. 
In this study, we set the $k$-hop radius as 3.

\subsection{Structure Learning for Discovering Spatial Correlations}

Observations have different types of variables, and NWP grid points also have their own node features. 
Thus, we apply fully-connected (FC) layers to unify their feature dimension sizes and embedding spaces of the observations and grid points. 
We first map input node features $X_t$ and $O_t$ into embedding space $\hat{X}_t, \hat{O}_t \in \mathbb{R}^{N \times d'}$ using FC layers $FC$ with weights. 
This can be formulated as: 
\begin{equation}
    \hat{X_t} = \textrm{FC}_X(X_t), \hat{O_t} = \textrm{FC}_O(O_t).
\end{equation}

To discover highly-correlated neighborhoods for each NWP grid point, we rank edges between spatially adjacent NWP grid points and observations. 
For each node $v_{i,t}$, we estimate a set of scores $e_{i,t} = \{e_{i,j,t}\}_j^N$ that quantify its relevance to all nodes $v_{j,t}\in V_t$, including itself. 
To generate differentiable edge samples $e_i$, we use the Gumbel-Softmax \cite{Jang_2017_gumbel}. 

To score each edge $e_{i,j,t} \in e_{i,t}$ for node $v_{i,t}$, we consider node features of $v_{i,t}$ and $v_{j,t}$ and their spatial distance. 
When $v_{i,t}$ and $v_{j,t}$ are NWP grid points, there feature correlations ${c}_{i,j}$ can be formulated as $c_{i,j,t} = \sigma(W_{c1} \hat{x}_{i,t} \lVert W_{c2} \hat{x}_{j,t})$, where $W_{c1}$ and $W_{c2}$ are learnable parameters for linear projection, and $\sigma(\cdot)$ indicates an activation function. 
In addition, we consider spatial distances ${d}_{i,j}$ by passing through a linear layer as $\hat{d}_{ij} = \sigma(W_d {d}_{i,j})$ with learnable parameters $W_d$. 
Finally, an edge $e_{i,j,t}$ can be assessed as $p_{i,j,t} = \textrm{FC}_p(c_{i,j,t} \lVert \hat{d}_{i,j})$.

Each element $p_{i,j,t} \in p_{i,t}$ indicates a correlation between node $v_{i,t}$ and $v_{j,t}$.
Then, we apply Gumbel-Softmax over the edge probabilities $p_{i,j,t}$, we generate differentiable samples $\hat{e}_{i,t} \in \mathbb{R}^N$ with Gumbel noise $g_{i,t}$. 
This can be formulated as: 
\begin{align}
    &\hat{e}_{i,t} = \bigg\{\frac{\textrm{exp}((log(p_{i,j,t})+ g_{i,t}) + \tau)}{\Sigma_j \textrm{exp}((log(p_{i,j,t})+ g_{i,t}) + \tau)}\bigg|\forall j \in N \bigg\}, \nonumber \\
    &g_{i,t} \sim \textrm{Gumbel}(0,1),
\end{align}
where $\tau$ is a temperature parameter controlling the interpolation between a continuous categorical density and a discrete one-hot categorical distribution.

%Based on $\hat{e}_{i,t}$, we can compose an augmented adjacency matrix $\hat{A}_t$ of $G_t$. 
%However, if we use 

%If we use $\hat{e}_{i,t}$ directly to compose an augmented adjacency matrix, this can cause structural information loss or 

We sample edges of $v_{i,t}$ based on $\hat{e}_{i,t}$ using a method proposed by
\citet{Zhang_2023}. 
For selecting top-$k$ edges, we use a flexible node degree $k$ by sampling the edge per node from a learned distribution rather than a fixed $k$.
Focusing on a single node as before, we approximate the distribution of node embeddings $\hat{x}_{i,t} \in \mathbb{R}^d$ (or $\hat{o}_{i,t}$) following a VAE-like approach.
We encode its mean $\mu_{i,t} \in \mathbb{R}^d$ and variance $\sigma_{i,t} \in \mathbb{R}^d$ using FC layers and then reparameterize with noise $\epsilon_{i,t}$ to obtain embedding $\tilde{z}_{i,t} \in \mathbb{R}^d$. 
This can be formulated as: 
\begin{align}
&\mu_{i,t} = \text{FC}_\mu(\hat{x}_{i,t}), \quad \sigma_{i,t} = \text{FC}_\sigma(\hat{x}_{i,t}), \nonumber\\
&\tilde{z}_{i,t} = \mu_{i,t} + \epsilon_{i,t} \sigma_{i,t}, \quad \epsilon_{i,t} \sim \mathcal{N}(0, 1). 
\end{align}
We use the same process for observation nodes with $\hat{o}_{i,t}$ instead of $\hat{x}_{i,t}$. 

We concatenate each embedding variable $\tilde{z}_{i,t}$ with the L1-norm of the edge samples $\lVert \hat{e}_{i,t} \rVert_1$ and decode it into a scalar $k_{i,t} \in \mathbb{R}$ with another FC layer, representing a continuous relaxation of the neighborhood size for node $v_{i,t}$. This can be formulated as: 
\begin{align}
    k_{i,t} = \text{FC}_k\left(\tilde{z}_{i,t} \lVert \lVert \hat{e}_{i,t} \rVert_1 \right), 
\end{align}
where $\lVert$ indicates the concatenation operator. 
Since $\lVert \hat{e}_{i,t} \rVert_1$ is the same as a summation of edge probabilities for each node, it can be understood as representing an initial estimation of the node degree which is then improved by combining with an embedding $z_{i,t}$ based entirely on the node's features. 
Using edge samples to estimate degrees of nodes links these representation spaces to the primary latent space of the node features $\hat{X_t}$ and $\hat{O_t}$.
Based on $\hat{e}_{i,t}$ and $k_{i,t}$, we can compose an augmented adjacency matrix $\hat{A}_t$ of $G_t$. 

\subsection{Spatial and Temporal Feature Extraction}

In terms of the graph encoder, we have not made significant modifications to the GCN, which merely propagates messages between 1-hop neighborhoods. 
This enables us to assess the contribution of spatial correlation discovery using structure learning to forecasting accuracy, compared to sophisticatedly designed graph encoders. 
The $l$-th GNN layer can be formulated as: 
\begin{align}
&H_t^{(l+1)} = \sigma \left( H_t^{(l)} \tilde{A}_t W^{(l)} \right),
\\ 
&\tilde{A}_t = D^{-\frac{1}{2}} \hat{\hat{A}}_t D^{-\frac{1}{2}}, \hat{\hat{A}}_t = \hat{A}_t + I, 
\end{align}
where $W^{(l)}$ refers to the weight matrix of the $l$-th layer. 
We set initial node representations $H_t^{(0)}$ based on $\hat{X_t}$ and $\hat{O_t}$. 

Meteorological phenomena have various spatial scales. 
We consider these multi-scale characteristics and avoid over-smoothing and over-squashing issues by employing skip connections for multi-layer readout. 
Target node representations from GNN layers are concatenated and passed through linear transformation. 
This can be formulated as: 
\begin{align}
h_{i,t} = W_r \left( h_{i,t}^{(0)} \Big\lVert h_{i,t}^{(1)} \Big\lVert \cdots \Big\lVert h_{i,t}^{(L)} \right). 
\end{align}

We use GRU to extract temporal features and aggregate node representations obtained at multiple time points from $t-m+1$ to $t$. 
This can be formulated as: 
\begin{align}
z_{i,t-m+1:t} = \text{GRU} \left(h_{i,t-m+1}, \cdots, h_{i,t-1}, h_{i,t} \right), 
\end{align}
where $\text{GRU}(\cdot)$ indicates a GRU layer, and
$z_{i,t-m+1:t}$ is the final target node representation.

\subsection{Atmospheric State Forecasting}

By passing the final node representations $z_{i,t-m+1:t}$ through a few FC layers, we conduct the atmospheric state forecasting, as: $\tilde{x}_{i,t+1} = \text{FC}_{Fine} (z_{i,t-m+1:t})$. 
However, since observations appear at only one time point and have heterogeneous meteorological variables, directly training the model on the atmospheric state forecasting can cause a lack of understanding of correlations between heterogeneous observations. 
Thus, we first pre-tain the model with node feature reconstruction of every node at every time point with L1 loss, as: 
\begin{align}
\mathcal{L}_{Pre}(v_i, t) 
&= \lvert \text{FC}_{Pre} (z_{i,t-m+1:t}) - {x}_{i,t} \rvert + \lambda \lVert \theta \rVert_2^2 \nonumber
\\
&= \lvert \tilde{x}_{i,t} - {x}_{i,t} \rvert + \lambda \lVert \theta \rVert_2^2.
\end{align}
We use the same process for observation nodes with ${o}_{i,t}$ instead of ${x}_{i,t}$. 
Then we fine-tune the model for atmospheric state forecasting at NWP grid points with L1 loss, as: 
\begin{align}
\mathcal{L}_{Fine}(v_i, t+1) 
&= \lvert \text{FC}_{Fine} (z_{i,t-m+1:t}) - {x}_{i,t+1} \rvert + \lambda \lVert \theta \rVert_2^2 \nonumber 
\\
&= \lvert \tilde{x}_{i,t+1} - x_{i,t+1} \rvert + \lambda \lVert \theta \rVert_2^2.
\end{align}
The implementation of the proposed method will be made publicly available on GitHub repository upon the publication of this paper.

\section{Experiments}

We validated the effectiveness of the proposed model by comparing it with the existing STGNN models with and without structure learning. 

\subsection{Experimental Setup}

\subsubsection{Dataset}
To conduct an empirical evaluation, we collected real-world observations and atmospheric state data from the Korean Peninsula and the surrounding regions (approximately $30^\circ$N–$50^\circ$N, $120^\circ$E–$140^\circ$E).
We collected data from 1 June 2021 to 30 June 2021 at 6-hour intervals.
Both atmospheric state data and observations were obtained from the Korea Meteorological Administration (KMA). 
The KIM Variational Data Assimilation System (KVAR) was used as the reference atmospheric state, and the KIM Observation Processing Package (KPOP) was used for observation preprocessing.
The observational dataset \cite{Kang2018} comprised a total of 11 satellite and ground-based platforms, including AIRCRAFT (U, V, T), GPSRO (bending angle, BA), SONDE (U, V, T, Q), AMV (brightness temperature, TB), AMSU-A (TB), AMSR2 (TB), ATMS (TB), CrIS (TB), GK2A (TB), IASI (TB), and MHS (TB).
These sources provided a diverse set of variables, including wind components (U and V), temperature (T), specific humidity (Q), brightness temperature (TB) and bending angle (BA), enabling a comprehensive evaluation of the proposed model’s ability to integrate heterogeneous meteorological observations.
Due to data-sharing restrictions of KMA, the datasets are not publicly available.

Atmospheric states were sampled from the 28th vertical level of the NWP grid, which corresponds to an average pressure of 500 hPa. %— a key layer for mid- and long-range weather forecasting.
The observational data included multiple instrument types, which were mapped to the 500 hPa pressure level.
For satellite retrievals that do not provide direct pressure information, brightness temperature was used as the observed variable by wavelength. 
The Jacobian with respect to the 500 hPa pressure level was then computed to represent the weighting of each satellite channel’s contribution at this pressure. 
This enabled a fair comparison to be made across observation types.
All observational data were quality controlled and preprocessed using KMA's operational pipelines.

\begin{table*}[th]
\setlength{\tabcolsep}{0.09em}
\caption{Performance comparison of the proposed model and baselines on weather prediction tasks.}
\label{tab:results}
\begin{center}
\begin{footnotesize}
\begin{sc}
  \begin{tabular}{ccccccccccccccccc}
    \toprule
    \multirow{2}{*}{Model} & & \multicolumn{3}{c}{U (m/s)}& &  \multicolumn{3}{c}{V (m/s)}& & \multicolumn{3}{c}{T (K)}& & \multicolumn{3}{c}{Q (kg/kg)}\\
    \cmidrule{3-5}
    \cmidrule{7-9}
    \cmidrule{11-13}
    \cmidrule{15-17}
    \multirow{2}{*}{} & & RMSE ($\downarrow$) & MAE ($\downarrow$) & $R^2$ ($\uparrow$) & & RMSE ($\downarrow$) & MAE ($\downarrow$) & $R^2$ ($\uparrow$) & & RMSE ($\downarrow$) & MAE ($\downarrow$) & $R^2$ ($\uparrow$)  & & RMSE ($\downarrow$) & MAE ($\downarrow$) & $R^2$ ($\uparrow$) \\
    \midrule
    DCRNN  & & 0.058$\pm$0.047& 0.048$\pm$0.045 & 0.901$\pm$0.062 
    &  & 0.052$\pm$0.053 & 0.044$\pm$0.055 & 0.884$\pm$0.055
    &  & 0.077$\pm$0.057& 0.068$\pm$0.052 & 0.792$\pm$0.073
    &  & 0.079$\pm$0.051 & 0.065$\pm$0.048 & 0.873$\pm$0.080\\
    STGCN  & & 0.057$\pm$0.051 & 0.049$\pm$0.044 & 0.892$\pm$0.071 
    &  & 0.051$\pm$0.055 & 0.042$\pm$0.052 & 0.885$\pm$0.059
    &  & 0.076$\pm$0.059 & 0.069$\pm$0.051 & 0.784$\pm$0.077
    &  & 0.077$\pm$0.056 & 0.066$\pm$0.049 & 0.872$\pm$0.081\\
    GWNet  & & \underline{0.052$\pm$0.038} & 0.044$\pm$0.026 & \underline{0.918$\pm$0.058}
    &  & \underline{0.043$\pm$0.033} & 0.039$\pm$0.033 & \underline{0.893$\pm$0.052}
    &  & \underline{0.070$\pm$0.044}& \underline{0.061$\pm$0.039} & \underline{0.804$\pm$0.069}
    &  & \underline{0.069$\pm$0.045} & \underline{0.060$\pm$0.038} & \underline{0.885$\pm$0.061}\\
    AGCRN  & & 0.054$\pm$0.033 & \underline{0.043$\pm$0.028} & 0.916$\pm$0.052 
    &  & 0.045$\pm$0.036 & \underline{0.037$\pm$0.031} & 0.891$\pm$0.054
    &  & 0.071$\pm$0.048 & 0.063$\pm$0.041 & 0.801$\pm$0.073
    &  & 0.072$\pm$0.049 & 0.062$\pm$0.039 & 0.868$\pm$0.069\\
    CloudNine & & 0.063$\pm$0.042 & 0.052$\pm$0.034 & 0.799$\pm$0.064 
    &  & 0.060$\pm$0.045 & 0.054$\pm$0.041 & 0.816$\pm$0.066
    &  & 0.074$\pm$0.054 & 0.066$\pm$0.049 & 0.759$\pm$0.087 
    &  & 0.078$\pm$0.056 & 0.067$\pm$0.047 & 0.784$\pm$0.082\\
    \midrule
    CloudNine-v2  
    & & \textbf{0.049$\pm$0.023} & \textbf{0.038$\pm$0.025} & \textbf{0.920$\pm$0.045} 
    &  & \textbf{0.040$\pm$0.022} & \textbf{0.033$\pm$0.021} & \textbf{0.904$\pm$0.044} 
    &  & \textbf{0.067$\pm$0.038} & \textbf{0.059$\pm$0.036} & \textbf{0.814$\pm$0.062} 
    &  & \textbf{0.068$\pm$0.033} & \textbf{0.058$\pm$0.031} & \textbf{0.881$\pm$0.048} \\
    \bottomrule
  \end{tabular}
\end{sc}
\end{footnotesize}
\end{center}
\label{tab:main}
\end{table*}

\begin{table*}[h]
\setlength{\tabcolsep}{0.09em}
\caption{
Node-level $R^2$ for four meteorological variables under low- and high-variability groups.
$|\Delta|$ indicates the absolute difference between the two accuracies ($|\Delta| = |\text{Low}-\text{High}|$).
Nodes were grouped into ‘low variability’ (bottom 25\%) and ‘high variability’ (top 25\%) using the variability index (VI).
%Nodes were categorized as “low variability” (bottom 25\%) or “high variability” (top 25\%) according to the VI.
}
\label{tab:results}
\begin{center}
\begin{footnotesize}
\begin{sc}
  \begin{tabular}{ccccccccccccccccc}
    \toprule
    \multirow{2}{*}{Model} & & \multicolumn{3}{c}{U (m/s)}& &  \multicolumn{3}{c}{V (m/s)}& & \multicolumn{3}{c}{T (K)}& & \multicolumn{3}{c}{Q (kg/kg)}\\
    \cline{3-5}
    \cline{7-9}
    \cline{11-13}
    \cline{15-17}
    \multirow{2}{*}{} & & Low & High & $|\Delta|$ && Low & High & $|\Delta|$ && Low & High & $|\Delta|$ && Low & High & $|\Delta|$ \\
\midrule
    DCRNN   &    & 0.825$\pm$0.058 & 0.411$\pm$0.073 & 0.414$\pm$0.017 
    && 0.854$\pm$0.041 & 0.496$\pm$0.078 & 0.358$\pm$0.037 
    && 0.549$\pm$0.069 & 0.321$\pm$0.078 & 0.228$\pm$0.016 
    && 0.789$\pm$0.080 & 0.445$\pm$0.091 & 0.344$\pm$0.011 \\
    STGCN  &    & 0.834$\pm$0.064 & 0.454$\pm$0.071 & 0.380$\pm$0.015 
    && 0.862$\pm$0.049 & 0.483$\pm$0.082 & 0.379$\pm$0.033 
    && 0.541$\pm$0.073 & 0.324$\pm$0.087 & 0.217$\pm$0.014 
    && 0.782$\pm$0.079 & 0.453$\pm$0.081 & 0.329$\pm$0.002 \\
    GWNet  &    & \underline{0.921$\pm$0.053} & 0.837$\pm$0.070 & 0.084$\pm$0.011 
    && \underline{0.890$\pm$0.042} & 0.793$\pm$0.067 & 0.097$\pm$0.024 
    && \underline{0.742$\pm$0.061} & \underline{0.678$\pm$0.077} & \underline{0.064$\pm$0.015} 
    && \underline{0.841$\pm$0.054} & \underline{0.744$\pm$0.070} & \underline{0.097$\pm$0.015} \\
    AGCRN   &   & 0.868$\pm$0.046 & \underline{0.844$\pm$0.053} & \underline{0.024$\pm$0.013} 
    && 0.841$\pm$0.033 & \underline{0.817$\pm$0.062} & \underline{0.024$\pm$0.029}
    && 0.645$\pm$0.067 & 0.549$\pm$0.081 & 0.096$\pm$0.014 
    && 0.832$\pm$0.068 & 0.637$\pm$0.073 & 0.195$\pm$0.006 \\
    CloudNine&  & 0.739$\pm$0.054 & 0.321$\pm$0.067 & 0.418$\pm$0.014 
    && 0.752$\pm$0.040 & 0.454$\pm$0.071 & 0.298$\pm$0.031 
    && 0.488$\pm$0.081 & 0.363$\pm$0.089 & 0.125$\pm$0.017 
    && 0.568$\pm$0.073 & 0.334$\pm$0.091 & 0.234$\pm$0.019 \\
    \midrule
    CloudNine-v2
               && \textbf{0.923$\pm$0.042} & \textbf{0.905$\pm$0.046} & \textbf{0.018$\pm$0.009}
               && \textbf{0.892$\pm$0.033} & \textbf{0.871$\pm$0.053} & \textbf{0.021$\pm$0.020}
               && \textbf{0.806$\pm$0.054} & \textbf{0.786$\pm$0.068} & \textbf{0.020$\pm$0.008}
               && \textbf{0.839$\pm$0.044} & \textbf{0.823$\pm$0.049} & \textbf{0.016$\pm$0.005} \\
    \bottomrule
  \end{tabular}
\end{sc} 
\end{footnotesize}
\end{center}
\end{table*}

\subsubsection{Baseline Models}

Although various STGNN models have been developed for general spatiotemporal forecasting, studies targeting meteorological prediction specifically and incorporating heterogeneous observations remain limited. 
CloudNine \cite{Jeon24MLST}, a GNN-based model for multi-source meteorological prediction, is one of the few existing approaches and was used as the primary baseline in our experiments.
However, it should be noted that CloudNine is not explicitly designed for time series prediction, but for multi-source feature integration in meteorological applications. 
To rigorously assess the effectiveness of our proposed method, we also compare it with several STGNN architectures that are widely adopted for spatiotemporal modelling in related fields.
We compare our proposed model with the following baselines:
\begin{itemize}
\item DCRNN \cite{Li2018} 
employs diffusion convolutional recurrent neural networks to model spatial and temporal dependencies in time series data over graphs, enabling robust multi-step forecasting.
\item STGCN \cite{Yan_stgnn} 
utilizes spatial-temporal graph convolutional layers to simultaneously capture local spatial dependencies and temporal patterns in dynamic graphs.
\item GWNet \cite{Wu_gwnet} 
learns adaptive graph structures via node-wise gating mechanisms and applies dilated causal convolutions for long-term time series forecasting.
\item AGCRN \cite{Bai_agcrn} 
introduces an adaptive graph convolution module that dynamically learns node embeddings and adjacency matrices, facilitating effective modeling of spatially heterogeneous relationships.
\item CloudNine \cite{Jeon24MLST} 
integrates multi-source meteorological data using deep learning, employing advanced architectures to enhance predictive accuracy for weather variables. 
\end{itemize}
We use the official implementations for all baselines and tune the hyperparameters on the validation set.
The implementation of CloudNine-v2 will be made publicly available in an open-source repository\footnote{https://github.com/higd963/CloudNine-v2} upon acceptance.

\subsubsection{Evaluation Protocol}
We adopt three widely used evaluation metrics to evaluate model performance: mean absolute error (MAE), root mean squared error (RMSE), and $R^2$ score. 
All datasets are split with a ratio of 6:2:2 into training, validation, and test sets.
We use 8 time steps of historical weather data to predict the next 1 time point.
Experiments are conducted on CPU and NVIDIA A100.

\subsection{Performance Analysis}

Table~\ref{tab:main} summarizes the performance of our proposed model and several state-of-the-art spatiotemporal GNN baselines.
Our proposed model (CloudNine-v2) consistently outperforms the best-performance baselines (e.g., GWNet and AGCRN) across all evaluation metrics and variables. 
%In particular, the proposed model achieves the highest $R^2$ for both the U and V wind components, indicating a more accurate prediction of the wind fields compared to the baseline methods. 
%This demonstrates the effectiveness of our dynamic graph structure learning, which enables the model to capture evolving spatial dependencies more effectively than approaches that rely on pre-defined graphs.
While several baselines achieve competitive results for certain variables, our method demonstrates relatively uniform and reliable accuracy across all four meteorological variables, including temperature (T) and specific humidity (Q).
This robustness suggests that integrating multi-source meteorological observations and adaptively modeling influence radii enhances generalizability of model across diverse weather variables.
Although AGCRN and GWNet also employs dynamic adjacency learning, its improvements are limited compared to our model. 
This suggests that performance gains are not merely a result of increasing architectural complexity, but rather stem from our explicit edge sampling strategy. 
By adaptively selecting the most relevant neighbors through Gumbel-Softmax sampling, the proposed model avoids over-smoothing and preserves meaningful local structures, leading more accurate forecasts.

\begin{figure}[t]
    \includegraphics[width=\columnwidth]{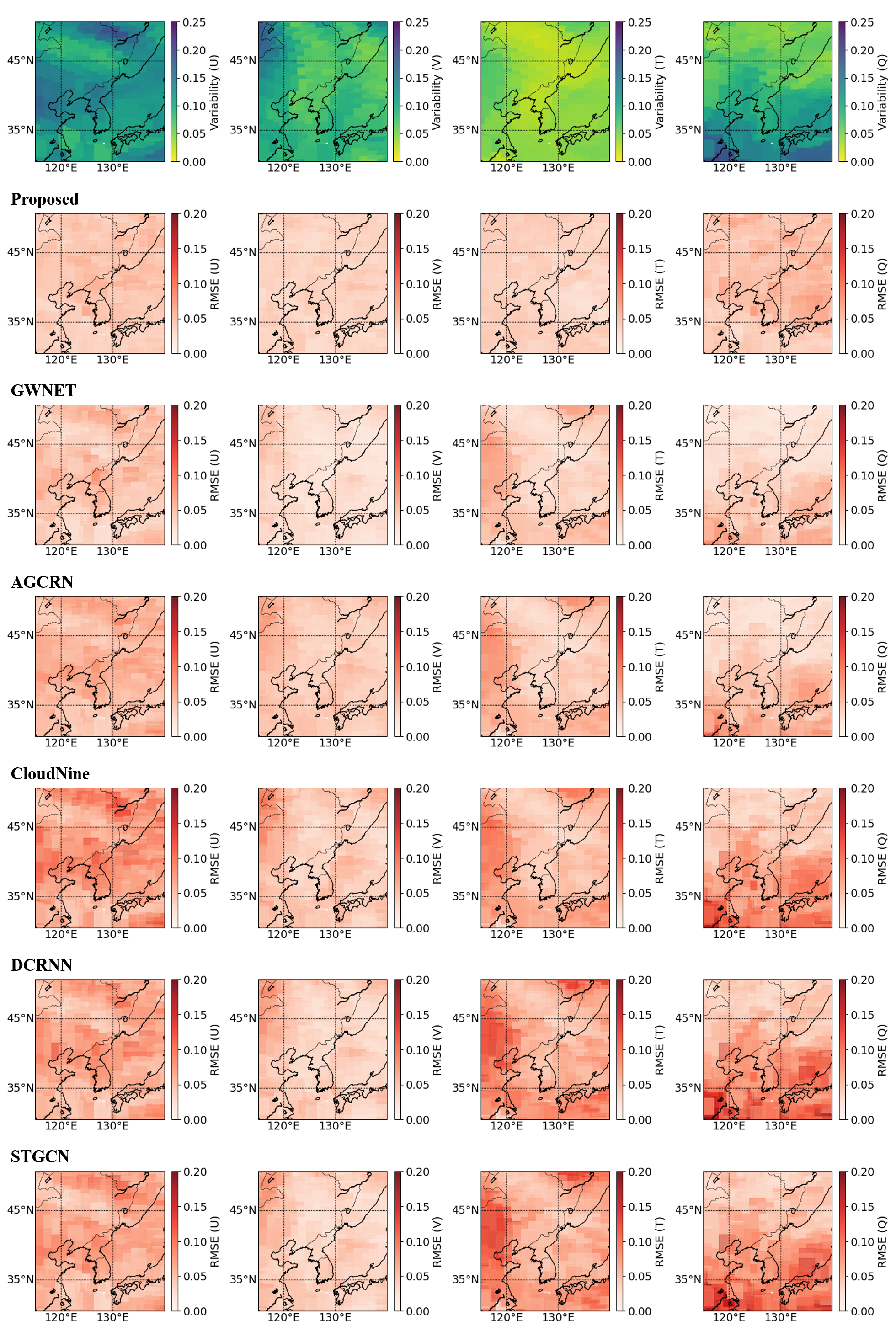}
    \caption{Node-level prediction accuracy for four meteorological variables, classified by low and high temporal variability groups.}
    \label{fig:results}
\end{figure}

Although our proposed model and GWNet demonstrate similar overall predictive performance in aggregate metrics, Table~\ref{tab:results} shows a significant difference in robustness in regions with atmospheric variability. 
The standard deviations showed in Tables~\ref{tab:main} are consistently smaller for our model across all variables. 
This indicates that the proposed method produces more stable and reliable predictions, regardless of the regional variability of the atmosphere. 
Also, the performance gap ($|\Delta|$) between low- and high-variability nodes is substantially smaller for our model than for GWNet, indicating superior stability in complex meteorological environments. 
As shown in Figure \ref{fig:results}, the robustness is particularly important in coastal and mountainous regions, where atmospheric variability is amplified by land-sea interactions and topographic effects, whereas inland areas typically exhibit lower variability and more consistent performance than ocean areas. 
%In contrast, inland areas with more stable atmospheric conditions typically exhibit lower temporal variability and, thus, less pronounced performance differences.
We attribute this advantage to our Gumbel-softmax-based top-$k$ edge selection, which adaptively reconfigures graph connectivity at each time step. 
Unlike GWNet's node-wise gating, which smoothly adjusts edge weights based on feature similarity, our approach makes context-sensitive updates to the set of active neighbors. 
Therefore, we can suggest that it enables rapid adaptation to abrupt changes in spatial dependencies and influence radii.
%By updating the set of active neighbors at each time step, our method more accurately reflects evolving patterns of local interactions, which are particularly pronounced in regions with high atmospheric variability. 
As a result, the model maintains stable predictive performance even in highly non-stationary atmospheric conditions.
%\begin{figure}[h]
%\begin{center}
%\centerline{\includegraphics[width=\columnwidth]{ICML2025/variability_node.png}}
%\caption{The performance comparison for four meteorological variables (U, V, T, and Q) across nodes with low and high temporal variability for the proposed model and three baseline methods.
%The proposed approach achieves higher, more stable accuracy in regions with high variability compared to all baselines.
%}
%\label{fig:var_node}
%\end{center}
%\end{figure}

\subsection{Ablation Studies}
Table~\ref{tab:ablation} presents an ablation study that assesses the impact of the model’s main components on its predictive performance.
When both adaptive adjacency ($\hat{A}$) and distance-based features ($dist$) are used, the model attains the highest $R^2$ scores across all variables. 
We can assume that the dynamic graph structure learning and spatial context information are important to weather forecasting.
Removing the distance-based features results in a significant decrease in $R^2$, especially for U and V, indicating the importance of spatial distance information in accurately capturing wind field dependencies.
The exclusion of adaptive adjacency, despite the presence of distance information, leads to moderate performance.
This suggests that, while distance information is helpful, explicitly learning dynamic graph structures enhances the model’s ability to represent complex spatial dependencies even more.
In conclusion, structure learning can cause excessive modifications of edges, which lead to structure information loss, noisy messages, loss of informative messages, and over-smoothing, and spatial distances can be used to effectively regulate the excessive modifications in STGNNs. 

\begin{table}[t]
\setlength{\tabcolsep}{0.38em}
\begin{small}
\begin{center}
\caption{Ablation study on the effect of main components (higher $R^2$ is better).}
\begin{tabular}{cccccc}
\toprule
%Task-spec Adj & Cross-View Fusion & Residual-Gated Fusion & Task 1 R² ↑ & Task 2 R² ↑ & Task 3 R² ↑ & Avg R² ↑ \\
$\hat{A}$ & $dist$ & U (m/s) & V (m/s) & T (K) & Q (kg/kg) \\
\midrule
\cmark & \cmark & \textbf{0.920$\pm$0.045} & \textbf{0.904$\pm$0.044} & \textbf{0.814$\pm$0.062} & \textbf{0.881$\pm$0.048} \\
\cmark & \xmark & 0.884$\pm$0.047 & 0.853$\pm$0.049 & 0.801$\pm$0.068 & 0.847$\pm$0.053 \\
\xmark & - & \underline{0.902$\pm$0.046} & \underline{0.881$\pm$0.048} & \underline{0.805$\pm$0.067} & \underline{0.873$\pm$0.051} \\
\bottomrule
\end{tabular}
\label{tab:ablation}
\end{center}
\end{small}
\end{table}

\subsection{Sensitivity Aanlysis}
\begin{figure}[ht]
\begin{center}
\centerline{\includegraphics[width=\columnwidth]{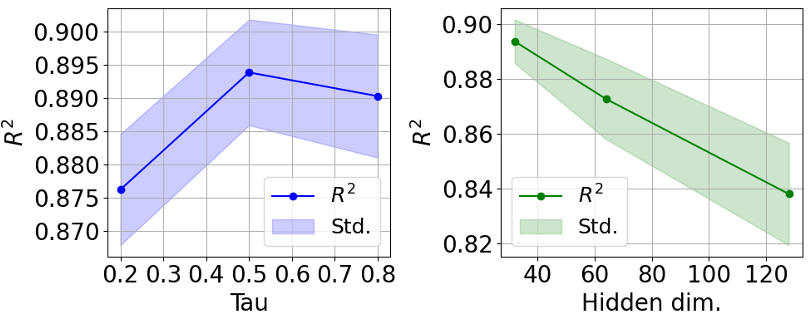}}
\caption{Sensitivity analysis of (left) the Gumbel-softmax temperature ($\tau$), and (right) the hidden dimension size, evaluated in terms of mean $R^2$ (solid line) and standard deviation (shaded area).}
\label{fig:sen}
\end{center}
\end{figure}

We conducted a sensitivity analysis to investigate the impact of the main hyperparameters on model performance, such as the Gumbel-softmax temperature ($\tau$), and the hidden dimension size, in Figure\ref{fig:sen}.

The moderate values of $\tau$ produce the highest accuracy, whereas lower and higher values lead to decreased accuracy and greater variability. 
These results imply that selecting an appropriate $\tau$ is essential for balancing exploration and exploitation in edge selection while maintaining robust model performance.
Also, we analyze the effect of the size of the hidden dimension. 
Increasing the hidden dimension beyond 32 results in a monotonic decrease in $R^2$ and an increase in standard deviation. 
This indicates that overly large hidden representations may cause overfitting or instability in the learning process. 
Therefore, our model achieves the best and most stable performance at $\tau = 0.5$, and hidden dimension = 32.

\section{Conclusion}

This study proposes a novel STGNN model that employs structure learning to improve the performance of atmospheric state estimation by discovering dynamically changing spatial correlations between NWP grid points and meteorological observations. 
The proposed model outperformed the existing STGNN models, and this underpins the effectiveness of structure learning for spatial correlations between meteorological data, which dynamically change. 
From the experimental results, we found that structure learning significantly contributes to the accuracy of atmospheric state estimation. 
Also, the proposed model slightly outperformed the SOTA STGNN models employing structure learning, while we merely use GCN as graph encoder. 
This improvement can be attributed to the fact that we regulate changes in graph structures by considering physical distances in structure learning. 
Importantly, the ability to integrate multi-source observations within a unified framework contributes to the robustness and generalizability of our model across diverse atmospheric variables.
In further research, we will focus on expanding the model to discover both spatial and temporal correlations.

\begin{acks}
This work was supported in part by the R\&D project “Development of a Next-Generation Data Assimilation System by the Korea Institute of Atmospheric Prediction System (KIAPS)”, funded by the Korea Meteorological Administration (KMA2020-02211) 
and
in part by the National Research Foundation of Korea (NRF) grant funded by the Korea government (MSIT) (No. 2022R1F1A1065516 and No. RS-2025-24523038) (O.-J.L.).
\end{acks}

\bibliographystyle{ACM-Reference-Format}
\bibliography{arXiv_2025} 

\end{document}